\pdfoutput=1
\documentclass{article}
\usepackage{amsmath,graphicx,mlspconf}
\usepackage[ruled,linesnumbered]{algorithm2e}
\usepackage{multicol}
\usepackage{multirow}
\usepackage{color,graphicx,amsthm,amssymb,amsmath,cite,xspace,setspace}
\usepackage[latin1]{inputenc}

\newcommand{\bmu}{\mbox{\boldmath   $\mu$}}

\title{IMAGE RESTORATION WITH LOCALLY SELECTED CLASS-ADAPTED MODELS}
\name{
\begin{tabular}{ccc}
Afonso M. Teodoro, & José M. Bioucas-Dias, & Mário A. T. Figueiredo\thanks{This work was partially supported by the (Portuguese) \textit{Fundação
		para a Ciência e Tecnologia} (FCT), grants UID/EEA/5008/2013 and
	BD/102715/2014.}
\end{tabular}
}
\address{
\begin{tabular}{c}
	Instituto de Telecomunicações and\\
	Instituto Superior Técnico, Universidade de Lisboa, \\
	Portugal
\end{tabular}
}

\DeclareMathOperator*{\argmin}{argmin}

\begin{document}
\ninept

\maketitle

\begin{abstract}
State-of-the-art algorithms for imaging inverse problems (namely deblurring and reconstruction) are typically iterative, involving a denoising operation as one of its steps. Using a state-of-the-art denoising method in this context is not trivial, and is the focus of current work. Recently, we have proposed to use a class-adapted denoiser (patch-based using Gaussian mixture models) in a so-called plug-and-play scheme, wherein a state-of-the-art denoiser is plugged into an iterative algorithm, leading to results that outperform the best general-purpose algorithms, when applied to an image of a known class (e.g. faces, text, brain MRI). 

In this paper, we extend that approach to handle situations where the image being processed is from one of a collection of possible classes or, more importantly, contains regions of different classes. More specifically, we propose a method to locally select one of a set of class-adapted Gaussian mixture patch priors, previously estimated from clean images of those classes. Our approach may be seen as simultaneously performing segmentation and restoration, thus contributing to bridging the gap between image restoration/reconstruction and analysis.

\end{abstract}
\begin{keywords}
Gaussian mixture models, image denoising, image deblurring, plug-and-play, class-adapted priors.
\end{keywords}
\section{Introduction}
\label{sec:intro}

Denoising is one of the oldest and most central problems in image processing, dating back to the early days of the field \cite{Lev}. Although it has been argued that the current best generic methods are very close to the theoretically maximum possible performance \cite{Chatterjee}, image denoising is still a very active area of research. Most, if not all, state-of-the-art methods belong to the patch-based family, i.e., they process the noisy image on a patch-by-patch fashion, relying on techniques such as non-local means \cite{Buades}, collaborative filtering \cite{dabov}, dictionary learning \cite{aharon}, or statistical models of the patches \cite{Teodoro2015}, \cite{yu}, \cite{ZoranWeiss}, \cite{zoran}. In more general imaging inverse problems, such as deblurring, computed tomography, or magnetic resonance imaging, to mention only a few classical examples, it may not be obvious how these patch-based techniques may be applied, and this has recently been a topic of interest \cite{danielyan}, \cite{Xu}, \cite{Yang}, \cite{ZoranWeiss}.
 
Whereas the bulk of the work in image restoration aims at developing methods of general applicability, there has been some recent work on developing class-specific methods \cite{luo}, \cite{Teodoro2016}. As the name suggests, these methods are tailored to perform very well on a certain class of images, for example, text, faces, or some type of medical images. Indeed, if we knew the class of the image being processed/reconstructed, it would be expectable that a targeted method could outperform a generic one, \textit{i.e.}, one that does not take into account the specific class of the image in hand.  Whereas in some cases, it is known that the image being estimated belongs to a certain class (e.g., brain MR or CT images, face images, fingerprints, text), in many situations this is not the case and there may even be regions of different classes present in the image (for example, a document image may contain text and one or more faces or some other type of images). This second type of scenario (where several classes may be present in a given image)  is the one addressed in this paper. Identifying the classes that are present at each location of an image can be interpreted as a form of image segmentation, thus the problem formulated and addressed in this paper may be seen as that of simultaneous image restoration and segmentation. 

Recently \cite{Teodoro2016}, we have proposed a method for class-adapted restoration/reconstruction, which builds upon the so-called \textit{plug-and-play} approach \cite{Venkatakrishnan}, by plugging a class-adapted denoiser based on Gaussian mixture models (GMM) into the iterations of an \textit{alternating direction method of multipliers} (ADMM) algorithm. Experiments reported in \cite{Teodoro2016} (both in deblurring and compressive imaging) have shown that the proposed method yields state-of-the-art results, when applied to  images known to contain text or a face, clearly outperforming the best generic techniques, such as IDD-BM3D \cite{danielyan}. 

In this paper, we extend the plug-and-play class-adapted approach to the scenario mentioned above: the image being restored is from an unknown class and/or may contain regions from different classes (e.g., text and faces, or text and natural images). As mentioned above, this may be seen as a method to perform simultaneous segmentation and restoration, by exploiting synergies between these two tasks. Notice that our main goal is not to obtain a good or meaningful segmentation, but to use the segmentation to allow class-specific models to be exploited at each location of the image, thus our focus is on the restoration performance of the method. Nevertheless, we believe that this type of approach may contribute to bridging the gap between low-level (restoration) and mid-level (segmentation) image processing/analysis.

The remaining sections of the paper are organized as follows. Section \ref{sec:tools} reviews the classical image restoration formulation and the basic tools upon which our method is built: ADMM, the plug-and-play scheme, and patch-based image denoising using GMM priors. The proposed method is described in Section \ref{sec:method}, while experimental results are reported in Section \ref{sec:results}. Finally, Section \ref{sec:conclusions} concludes the paper and gives pointers for future work.

\section{Formulation and Tools}\label{sec:tools}
\subsection{Problem Formulation}
The classical formulation of image reconstruction/restoration problems is
\begin{equation}
\textbf{y} = \textbf{Ax} + \textbf{n}, \label{eq:1}
\end{equation} 
where $\textbf{y} \in \mathbb{R}^m$, $\textbf{x} \in \mathbb{R}^n$ are the observed data and the underlying clean vectorized image, respectively, $\textbf{A} \in \mathbb{R}^{m \times n}$ is the observation matrix, and $\textbf{n}$ is noise. For simplicity, we  assume that the noise is Gaussian, with zero mean and known variance $\sigma^2$. In denoising, the operator $\bf A$ is the identity, but in more general cases it is non-invertible or ill-conditioned,  making \eqref{eq:1} an ill-posed inverse problem. Arguably, the most common approach to handle ill-posed problems of the form \eqref{eq:1} is to seek a {\it maximum a posteriori} (MAP) estimate,
\begin{equation}
\hat{\textbf{x}} = \underset{\textbf{x}}{\argmin} \, \, \frac{1}{2\sigma^2} \Vert \textbf{Ax} - \textbf{y} \Vert_2^2 +  \phi(\textbf{x}), \label{eq:analysis}
\end{equation}
where $\frac{1}{2\sigma^2} \Vert \textbf{Ax} - \textbf{y} \Vert_2^2 = -\log p({\bf y}|{\bf x}) + A$ (with $A$ an irrelevant constant) is  the negative log-likelihood and $\phi(\textbf{x}) \propto -\log p(\bf x)$ is the negative log-prior. Some recent work has also been devoted to obtaining \textit{minimum mean squared error} (MMSE) estimates, rather than MAP, which corresponds to the posterior expectation $\mathbb{E}[\bf x | y]$ \cite{Kazerouni}, \cite{Louchet}.
The criterion in \eqref{eq:analysis} may also be given a non-probabilistic interpretation, and seen as a regularization approach.

\subsection{ADMM and Plug-and-Play}\label{sec:admm}
In recent years, variable splitting algorithms, such as ADMM, have  received a lot of attention, in particular in the image processing and machine learning communities \cite{boyd}. A notable feature of ADMM, as applied to \eqref{eq:analysis}, is that it separates the handling of the data term (log-likelihood)  from that of the prior/regularizer. The standard instantiation of ADMM to tackle \eqref{eq:analysis} consists in the cyclic application of the following steps \cite{afonso}, \cite{Almeida2013}: 
\begin{align}
\textbf{x}^{k+1} & := \underset{\textbf{x}}{\argmin} \, \, \frac{1}{2} \Vert \textbf{Ax} - \textbf{y} \Vert_2^2+ \frac{\mu}{2} \Vert \textbf{x} - \textbf{v}^k - \textbf{d}^k \Vert_2^2, \label{eq:x}\\
\textbf{v}^{k+1} & := \underset{\textbf{v}}{\argmin} \, \, \phi(\textbf{v}) + \frac{\mu}{2} \Vert \textbf{x}^{k+1} - \textbf{v} - \textbf{d}^k \Vert_2^2, \label{eq:v}\\
\textbf{d}^{k+1} & := \textbf{d}^k - (\textbf{x}^{k+1} - \textbf{v}^{k+1}).\label{eq:d}
\end{align}

Problem \eqref{eq:x} is quadratic and has a closed-form solution, which requires solving a linear system (equivalently, inverting a matrix):
\begin{equation}
\textbf{x}^{k+1} = (\textbf{A}^T\textbf{A} + \mu \, \textbf{I})^{-1}\bigl((\textbf{A}^T\textbf{y} + \mu \, (\textbf{v}^k+\textbf{d}^k)\bigr). \label{eq:inv}
\end{equation}
Although this is sometimes seen as an obstacle (motivating, for example, the introduction of linearized versions of ADMM \cite{LADMM}), there are cases in which this inversion can be computed very efficiently using fast transforms (namely the FFT) \cite{afonso}, \cite{Almeida2013}, and in those cases ADMM exhibits state-of-the-art speed. Image deconvolution (with periodic \cite{afonso} or other boundary conditions \cite{Almeida2013}) and inpainting are two of those cases where ADMM excels.   

The update expression in \eqref{eq:v} corresponds to the so-called {\it Moreau proximity operator} (MPO) of $\phi$ \cite{Bauschke}, computed at $\textbf{x}^{k+1} - \textbf{d}^k$. Recall that the MPO of $\phi:\mathbb{R}\rightarrow \bar{\mathbb{R}}  = \mathbb{R}\cup \{+\infty\}$ is defined as 
\[
\mbox{prox}_{\phi}({\bf z}) = \underset{\textbf{x}}{\argmin} \frac{1}{2}\| {\bf x - z}\|_2^2 + \phi({\bf x}),
\] 
and can be seen as the MAP solution of a denoising problem, where the argument of $\mbox{prox}_{\phi}$ is the noisy data, the noise is Gaussian i.i.d. with unit variance, and the prior is $p({\bf x}) \propto \exp(-\phi(\bf x))$.

In recent work, it has been suggested that the denoiser corresponding to the MPO in \eqref{eq:v} can be replaced with an off-the-shelf  state-of-the-art denoising algorithm, such as BM3D \cite{dabov}; that approach has been termed {\it plug-and-play} \cite{Venkatakrishnan}. Since, in general, a denoising algorithm may not correspond necessarily to the MPO of a convex regularizer/log-prior, convergence of the resulting algorithm requires further analysis; initial results have been presented in \cite{Venkatakrishnan}.

\subsection{Patch-Based Denoising with GMM Priors}
It has been shown that a simple GMM, estimated from a collection of clean images, is a remarkably effective patch prior \cite{ZoranWeiss}. More recently, we have shown that excellent denoising performance is also obtained if this GMM is estimated directly from the patches of the noisy image to be denoised, and that the corresponding expectation-maximization (EM) algorithm is a simple modification of the one for estimating the GMM from noiseless patches \cite{Teodoro2015}. Moreover, since the prior is a GMM, it is easy to compute the conditional expectation of each patch given its noisy version, which is the optimal MMSE estimate (in contrast with \cite{ZoranWeiss}, where a MAP estimate is used). Letting ${\bf x}_i$ and ${\bf y}_i$ denote an arbitrary patch of images $\bf x$ and $\bf y$ (unknown clean image and noisy one, respectively), the probabilistic model for patch denoising is
\begin{eqnarray}
p({\bf x}_i) & = & \sum_{m=1}^K \alpha_m \; \mathcal{N}({\bf x}_i;\bmu_m,{\bf C}_m)\\
p({\bf y}_i|{\bf x}_i) & = & \mathcal{N}({\bf y}_i ; {\bf x}_i,\sigma^2\, {\bf I})
\end{eqnarray}
where $\mathcal{N}(\cdot ; \bmu,{\bf C})$ denotes a Gaussian probability density function of mean $\bmu$ and covariance $\bf C$.
The resulting  MMSE estimate of ${\bf x}_i$ is given by
\begin{equation}
\hat{\bf x}_i = \sum_{m=1}^K \beta_m({\bf y}_i ) \; {\bf v}_m({\bf y}_i),\label{MMSE}
\end{equation}
where
\begin{equation}
{\bf v}_m({\bf y}_i) = \Bigl( \sigma^2\, {\bf C}_m + {\bf I} \Bigr)^{-1}\Bigl( \sigma^2\, {\bf C}_m^{-1}\bmu_m +  {\bf y}_i\Bigr),
\end{equation}
and 
\begin{equation}
\beta_m({\bf y}_i) = \frac{\alpha_m \; \mathcal{N}({\bf y}_i ; \bmu_m, {\bf C}_m + \sigma^2 \, {\bf I})
	 }{\sum_{j=1}^K \alpha_j \; \mathcal{N}({\bf y}_i ; \bmu_j, {\bf C}_m + \sigma^2 \, {\bf I})}.
\end{equation}
Notice that $\beta_m({\bf y}_i)$ is simply the posterior probability that the $i$-th patch was generated by the $m$-th GMM component, whereas ${\bf v}_m({\bf y}_i)$ is the conditional MMSE (and MAP) estimate of ${\bf x}_i$, if we knew that it had been generated by the $m$-th GMM component.

The final denoised image is assembled by putting the patch estimates back in their locations. Since the patches overlap, there are several estimates of each pixel, which are usually combined by straight averaging. In \cite{Teodoro2015}, we proposed to use the optimal weighted averaging, where the weights are the inverses of the posterior variances, which can also be computed in closed-form. The use of weights based on the posterior variance of the patch estimates was also used in \cite{Chatterjee}, but with a single Gaussian per patch.

Recently \cite{Teodoro2016}, we have proposed to use GMM patch-based denoising in a plug-and-play approach, to deblur images of specific classes. For that purpose, the GMM is estimated from a collection of clean images of the class of interest, and the denoising step \eqref{eq:v} of the ADMM algorithm is replaced with the GMM-patch-based method described above. Experiments reported in \cite{Teodoro2016} show that the  method produces excellent results in deconvolving text and face images, outperforming the generic state-of-the-art method IDD-BM3D \cite{danielyan}. 

\section{Proposed Method}\label{sec:method}
In this paper, we extend the method proposed in \cite{Teodoro2016} to handle images of unknown classes or even containing regions from different classes. Rather than a single class-adapted GMM, estimated from a collection of clean images from that class, consider $C$ classes, each of which modelled by a GMM,
\begin{equation}
p({\bf x}_i | c_i ) =  \sum_{m=1}^{K^{(c_i)}} \alpha_m^{(c_i)} \; \mathcal{N}({\bf x}_i;\bmu_m^{(c_i)},{\bf C}_m^{(c_i)}),
\end{equation}
where $c_i \in \{1,...,C\}$ is the class label of the $i$-the patch, and $C$ the total number of classes. To estimate the $i$-th patch, we begin by classifying it into one of the classes, and then use the corresponding GMM to obtain an MMSE estimate of that patch (given by \eqref{MMSE}), conditioned on its noisy version. We emphasize that this process is repeated until some stopping criterion is met.

As mentioned in Section 1, classifying each patch into one of the classes can be seen as performing image segmentation. However, we stress again that segmentation is not the main goal of the proposed approach, thus we will only focus on its performance in terms of denoising and deblurring. To classify the patches, we consider two alternatives: 
\begin{itemize}
\item Simply classifying each patch independently using the maximum-likelihood criterion, 
\begin{equation}
\hat{c}_i = \arg\max_{c\in\{1,...C\}}  p({\bf x}_i | c ).\label{eq:ml}
\end{equation} 
\item Jointly classifying all the patches under a Markov random field prior $p({\bf c})$ (where ${\bf c}$ denotes the field of all the patch class labels), more specifically a Potts prior \cite{boykov98}, 
\begin{equation}
\hat{\bf c} = \arg\max_{c\in\{1,...C\}} \log p({\bf c}) + \sum_i \log p({\bf x}_i | c_i ). \label{MRF}
\end{equation}
To solve \eqref{MRF} we use the $\alpha$-expansion graph-cut algorithm proposed in \cite{boykov}. For more details about MRF priors for image segmentation, see \cite{boykov}, \cite{boykov98}. 
\end{itemize}

In summary, the multi-class denoiser described in the two preceding paragraphs is used in the plug-and-play approach described in Subsection \ref{sec:admm} (see Algorithm~\ref{alg:deblur}).

\SetKwInOut{Parameter}{Parameters}

\begin{algorithm}
	\SetAlgoLined
	\KwIn{Blurred image, blur kernel, generic GMM, class-specific GMMs\;}
	\KwOut{Deblurred image, patch classification\;}
	\Parameter{Patch size, $\mu$\;}
	\Repeat{stopping criterion}{
		Solve \eqref{eq:x}\;
		Extract deblurred image patches\;
		Classify each patch - \eqref{eq:ml} or \eqref{MRF}\;
		Solve \eqref{eq:v}
		
		\hspace{1.5em} Denoise each patch using class-adapted GMM \eqref{MMSE}\;
		Combine all estimates of each image pixel\;
		Update dual variable \eqref{eq:d}\;
	}
	\caption{Proposed ADMM-GMM with classification \label{alg:deblur}}
\end{algorithm}

%
%
%

\section{Experimental Results}
\label{sec:results}

We start by presenting some results on image denoising. Table~\ref{tab:den1} compares the results obtained with the denoising algorithm, with and without classification of the patches. As a baseline, we also present the results of a state-of-the-art denoising algorithm, BM3D \cite{dabov} (with default parameters). In every run, the patch size was set to 8 by 8, and we trained a GMM with 20 components on the noisy patches, using the approach described in \cite{Teodoro2015}. Furthermore, the following classes were considered: \textit{text, faces, brain MRI, fingerprints}, and \textit{generic}. All of the external GMMs, also with 20 components each, were trained with samples from the corresponding class, except the \textit{generic} GMM, which was trained using random images from the Berkeley dataset for image segmentation (BSDS300) \cite{berkeley}. 

\begin{table*}
\begin{center}
\resizebox{\textwidth}{!}{
\begin{tabular}{ c||c|c|c|c||c|c|c|c||c|c|c|c}
\hline
\multirow{2}{*}{$\sigma$}& \multicolumn{4}{|c||}{Cameraman} & \multicolumn{4}{c||}{House} &  \multicolumn{4}{c}{Text}\\
\cline{2-13}
 & \multicolumn{1}{|c|}{BM3D} &  GMM & GMM (C) & GMM ($\alpha$) & \multicolumn{1}{|c|}{BM3D}  & GMM & GMM (C) & GMM ($\alpha$)& \multicolumn{1}{|c|}{BM3D}  & GMM & GMM (C) & GMM ($\alpha$) \\ \hline

5  & 38.29  & 38.37 & \textbf{38.39} & \textit{38.38} & 39.83 & \textbf{39.89} & \textit{39.88} & 39.87 & 39.01  & 40.15  & \textbf{40.50} &  \textit{40.38} \\ \hline
15  & 31.91 & \textit{31.94} & \textbf{32.00} & \textbf{32.00} & \textbf{34.94} & \textit{34.78} & 34.77 & 34.76 & 30.35 &  31.56 &\textit{ 31.79 }& \textbf{31.84} \\ \hline
30  & \textbf{28.64} & 28.46 &  \textit{28.51} & 28.50 & \textbf{32.09} & 31.83 & \textit{31.84} & 31.83 & 25.04  & 26.25 & \textit{26.61} & \textbf{26.70} \\ \hline
50 & 26.12 & \textit{26.14} & \textbf{26.17} & \textbf{26.17} & \textbf{29.69} & 29.42 & \textit{29.43} & 29.38 & 20.45 & 22.65 & \textbf{22.96} & \textit{22.81} \\  \hline
100 & \textbf{23.07} & 22.97 & \textit{22.99} & \textit{22.99} & 25.87 & \textbf{25.90} & \textbf{25.90} & \textit{25.89} & 16.19 & \textit{17.75} & \textbf{17.95} & 17.43 \\ \hline \hline
\multirow{2}{*}{$\sigma$}& \multicolumn{4}{|c||}{Face} & \multicolumn{4}{c||}{Cameraman + Text} &  \multicolumn{4}{c}{Face + Text}\\
\cline{2-13}
 & \multicolumn{1}{|c|}{BM3D} &  GMM & GMM (C) & GMM ($\alpha$) & \multicolumn{1}{|c|}{BM3D}  & GMM & GMM (C) & GMM ($\alpha$) & \multicolumn{1}{|c|}{BM3D}  & GMM & GMM (C)  & GMM ($\alpha$) \\ \hline

5 & 39.70  &  39.81 & \textbf{39.97} &\textit{ 39.84} & 38.46  & 38.59 & \textbf{38.70} & \textit{38.68} & 35.39 & \textbf{36.01} & 35.82 & \textit{35.86} \\ \hline
15 & \textit{33.38} &  33.34 & \textbf{33.47} & 33.14 & 31.58 & 31.77 & 31.91 & \textbf{31.95} & 28.78 & \textbf{29.50} &\textit{ 29.47 }& \textit{29.47} \\ \hline
30 & \textbf{29.36}  & 29.20 & \textit{29.24} & 29.15 & 27.80 & 27.84 & \textbf{28.06} & 28.04 & 23.82 & \textit{24.83}  & 24.82 & \textbf{24.86}\\ \hline
50 & \textbf{26.79} & 26.19 & \textit{26.31} & 26.30 & 24.52 & 25.10 & \textbf{25.37} & 25.34 & 19.96 & 21.58 & \textbf{21.63} &\textit{ 21.59} \\  \hline
100 & \textbf{22.96} & 22.33 & 22.16 & \textit{22.45} & 21.05 & \textbf{21.59 }& 21.14 & 21.18 & 16.46 & 17.36 & \textbf{17.44} & \textit{17.39} \\ \hline
\end{tabular}
}
\end{center}
\caption{PSNR on image denoising - Methods: BM3D \cite{dabov}; GMM denoiser \cite{Teodoro2015}; GMM denoiser with patch classification (C); GMM denoiser with $\alpha$-expansion ($\alpha$).\label{tab:den1}}
\end{table*}

We conclude that the proposed modification does not have a significant impact on the denoising performance. Arguably, this is due to the fact that, in pure denoising, it is possible to estimate a GMM from the noisy image itself \cite{Teodoro2015}. The resulting model is thus better adapted to the input image than if it would be trained from a different set of images, even if these images are from the same class. 

\begin{figure}
\resizebox{0.5\textwidth}{!}{
\begin{tabular}{ccc}
\includegraphics[width=.22\textwidth]{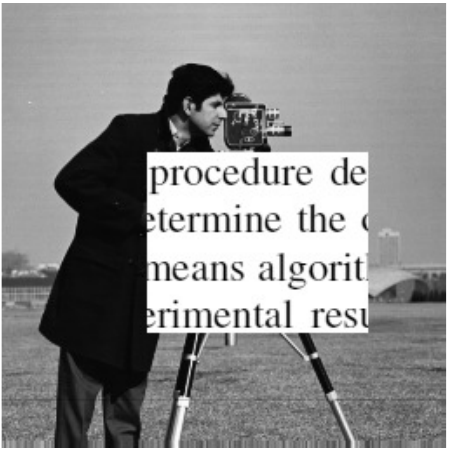}&
\includegraphics[width=.22\textwidth]{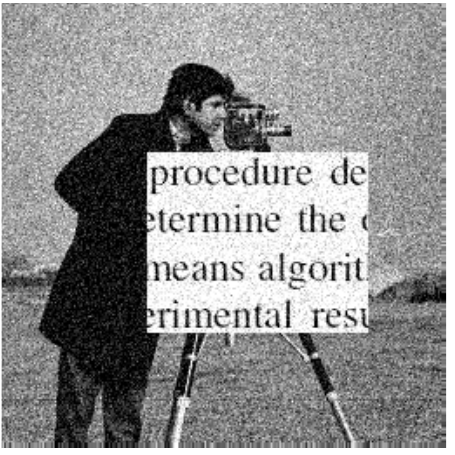}&
\includegraphics[width=.22\textwidth]{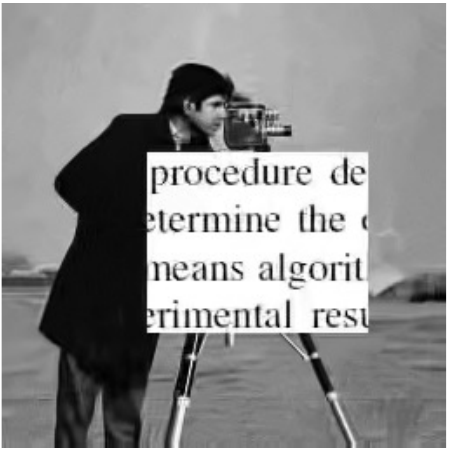}\\
(a)&(b)&(c)\\
\includegraphics[width=.22\textwidth]{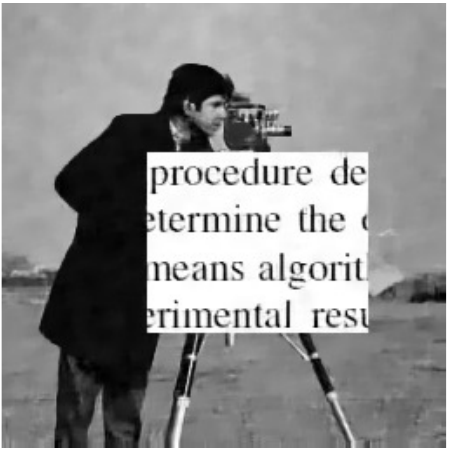}&
\includegraphics[width=.22\textwidth]{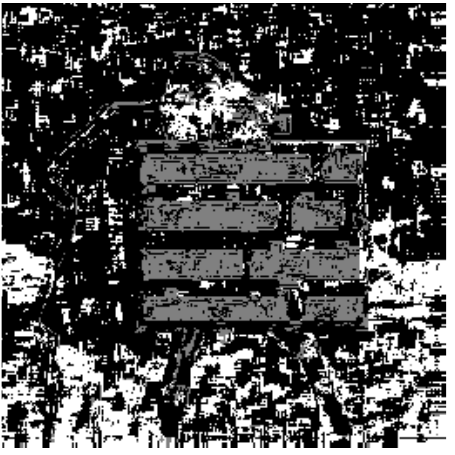}&
\includegraphics[width=.22\textwidth]{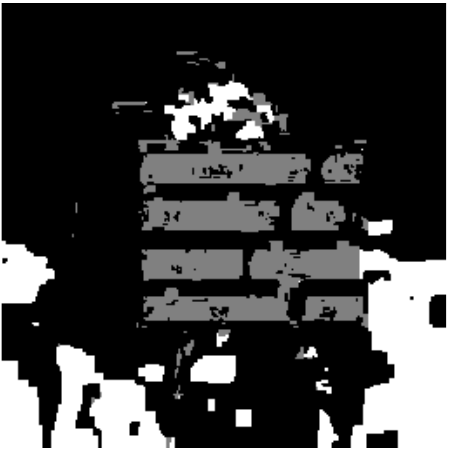}\\
(d) & (e)& (f)\\
\end{tabular}}
\caption{Denoising: (a) input image Cameraman + Text; (b) noisy image ($\sigma = 30$); (c) denoised with BM3D \cite{dabov} (PSNR = 27.80dB); (d) denoised with GMM with $\alpha$-expansion (PSNR = 28.10dB); (e) ML patch classification (f) patch classification with $\alpha$-expansion. }
\label{fig:den1}
\end{figure}

Figure~\ref{fig:den1} illustrates the difference in the patch labelling, if we consider classifying each patch independently via the maximum-likelihood (ML) criterion or using the $\alpha$-expansion graph-cut algorithm. While we do not expect the latter labelling to perform better in terms of denoising or deblurring, it is more meaningful from a segmentation viewpoint, since it is exhibits higher spatial coherence. In this example, we used three different models: one trained from the noisy image itself (black), one targeted to text images (grey), and one trained with \textit{generic} images (white).


In our deblurring experiments, we considered the same classes, where each GMM (with 20 components) was trained using $6\times 6$ patches. Table~\ref{tab:deb1} shows the results of ADMM-GMM algorithm, with and without classification, for all blur kernels in \cite{danielyan}. As a benchmark, we also present the results obtained with state-of-the-art IDD-BM3D \cite{danielyan} (with default parameters). In this set up, we assumed that the class of the input image is unknown a priori, even in the text or face images so, when no classification is done, the algorithm uses only the \textit{generic} GMM. Otherwise, we let the algorithm decide which class should be used for each patch, which explains the discrepancy that exists relative to the results reported in \cite{Teodoro2016}. Furthermore, after 100 iterations of the algorithm, we switched the \textit{generic} GMM with another GMM trained from the deblurred patches, which we assume to be reasonable estimates of the clean patches by then.

\begin{table*}
	\begin{center}
		\resizebox{\textwidth}{!}{
			\begin{tabular}{c||c|c|c|c|c|c||c|c|c|c|c|c}
				\hline 
				& \multicolumn{6}{c||}{Cameraman}                & \multicolumn{6}{c}{House}                    \\ \hline
				Experiment & 1     & 2     & 3     & 4     & 5     & 6     & 1     & 2     & 3     & 4     & 5     & 6     \\ \hline
				BSNR       & 31.87 & 25.85 & 40.00 & 18.53 & 29.19 & 17.76 & 29.16 & 23.14 & 40.00 & 15.99 & 26.61 & 15.15 \\ \hline
				Input PSNR & 22.23 & 22.16 & 20.76 & 24.62 & 23.36 & 29.82 & 25.61 & 25.46 & 24.11 & 28.06 & 27.81 & 29.98 \\ \hline \hline
				IDD-BM3D   &   \textbf{8.85}    &  \textbf{ 7.12}    &   \textbf{10.45}    &   \textbf{3.98}    &   \textbf{4.31}    &   \textbf{4.89}    &   \textbf{9.95}    &   \textbf{8.55 }   &   \textbf{12.89}    & \textbf{  5.79 }   &   \textbf{5.74}    & \textbf{ 7.13}   \\ \hline
				ADMM-GMM   &   8.39    &   6.36  &  9.80    &   3.47   &   4.16   &   \textit{4.88}   &   \textit{9.66}    &    \textit{8.22}   &   \textit{ 12.43}   &\textit{  5.50 }  &   \textit{5.42}    &   6.82  \\ \hline
				ADMM-GMM   (C) &   \textit{8.54}    &  \textit{6.44}  &   \textit{9.82}    &   \textit{3.49 }  &   \textit{4.20  } &  4.79 &   9.55    &  8.06   &   12.24 &  5.26  &  5.25   &   6.75 \\ \hline 
				ADMM-GMM   ($\alpha$) &   8.49    &    6.36  &   9.76    &   3.51  &    4.19   &   4.82   &   9.62    &   7.90   &    12.28   &  5.10   &   5.23    &   \textit{6.83}  \\ \hline\hline
				& \multicolumn{6}{c||}{Text}                & \multicolumn{6}{c}{Face}                    \\ \hline
				Experiment & 1     & 2     & 3     & 4     & 5     & 6     & 1     & 2     & 3     & 4     & 5     & 6     \\ \hline
				BSNR       & 26.07 & 20.05 & 40.00 & 15.95 & 24.78 & 18.11 & 28.28 & 22.26 & 40.00 & 15.89 & 26.22 & 15.37 \\ \hline
				Input PSNR & 14.14 & 14.13 & 12.13 & 16.83 & 14.48 & 28.73 & 25.61 & 22.54 & 20.71 & 26.49 & 24.79 & 30.03 \\ \hline \hline
				IDD-BM3D   &    11.97   &    8.91   &     16.29  &    5.88   &    6.81   &    4.87   &    13.66   &   11.16    &  14.96     &    7.31   &    10.33   &    6.18   \\ \hline
				ADMM-GMM   &   15.28   &    11.52  &   20.84    &   8.65   &    10.56  &   5.74   &   14.51    &   11.95    &    16.54   &  8.29   &   10.58    &   6.05  \\ \hline
				ADMM-GMM   (C) &   \textbf{15.69}    &  \textbf{12.06}  &   \textbf{21.58 }   &  \textbf{9.07}  &   \textbf{10.93}   &   \textbf{6.48}   &   \textbf{14.69}   &   \textit{12.23}   &    \textbf{17.01}   &  \textbf{8.38 } &    \textbf{11.12}   &   \textbf{6.94}  \\ \hline 
				ADMM-GMM   ($\alpha$) &   \textit{15.64}    &   \textit{ 12.00}  &   \textit{21.37}    &   \textit{9.06}   &    \textit{10.86}   &   \textit{6.33}   &  \textit{ 14.58 }   &   \textbf{12.59}    &    \textit{16.56}   & \textit{ 8.32} &   \textit{10.78}    & \textit{  6.88}  \\ \hline\hline
				& \multicolumn{6}{c||}{Cameraman + Text}                & \multicolumn{6}{c}{Face + Text}                    \\ \hline
				Experiment & 1     & 2     & 3     & 4     & 5     & 6     & 1     & 2     & 3     & 4     & 5     & 6     \\ \hline
				BSNR       & 32.98 & 26.95 & 40.00 & 19.71 & 30.29 & 19.18 & 23.51 & 22.26 & 40.00 & 17.30 & 27.48 & 18.14 \\ \hline
				Input PSNR & 18.87 & 18.84 & 17.15 & 21.61 & 19.65 & 29.59 & 15.38  & 22.54 & 14.52 & 16.84 & 15.72 & 28.49  \\ \hline \hline
				IDD-BM3D   &    10.83   &    8.71   &     12.39  &    5.40   &    6.27   &    4.89   &    9.71  &   6.57    &  11.24     &    2.43   &    1.79   &    \textbf{4.28}  \\ \hline
				ADMM-GMM   &   11.41   &    9.11  &   12.98    &   6.25  &    7.74  &   5.21   &   \textit{10.71}    &   7.55 &    12.56   &  3.97   &   2.88    &   4.10  \\ \hline
				ADMM-GMM   (C) &   \textbf{11.55}   & \textbf{9.32}  &   \textbf{13.25 }  &  \textbf{6.29}  &  \textbf{7.86 } &   \textbf{5.26}  &   \textbf{10.98 }  &   \textbf{7.79}   &    \textbf{12.59}  &  \textit{4.13}  &    \textbf{2.93}   &  \textit{4.21 } \\ \hline 
				ADMM-GMM   ($\alpha$) &   \textit{11.51}    & \textit{9.29}  &   \textit{13.21}   &  \textit{6.28 } &  \textit{7.75}  &  \textit{5.24}   & \textbf{ 10.98}  &  \textit{ 7.77}    &    \textit{12.58}   & \textbf{4.14} &  \textit{2.91}   & 4.18  \\ \hline
			\end{tabular}
		}
	\end{center}
	\caption{ISNR on image deblurring - Methods: IDD-BM3D \cite{danielyan}; ADMM with GMM prior \cite{Teodoro2016}; ADMM with GMM prior and patch classification (C); ADMM with GMM prior and $\alpha$-expansion ($\alpha$).\label{tab:deb1}}
\end{table*}

Since we cannot learn a GMM from the blurred input image, the improvement that is achieved with the proposed method is much more visible when, in fact, the input image contains one or more of the considered classes. On the one hand, regarding the Cameraman and House images (from the generic class), we observe that the proposed scheme performs worse than IDD-BM3D ($0.76$dB in the worst-case scenario), while using classification may or may not improve the performance. On the other hand, for the remaining examples, ADMM-GMM achieves better results than the generic IDD-BM3D. Although the parameters of IDD-BM3D were not tuned for these examples, we emphasize that when comparing ADMM-GMM with and without classification, the classification scheme that was proposed in this paper consistently improves the results on the images that contain one or more of the considered classes.

\begin{figure*}
	\centering
\resizebox{\textwidth}{!}{
\begin{tabular}{cccc}
\includegraphics[width=\textwidth]{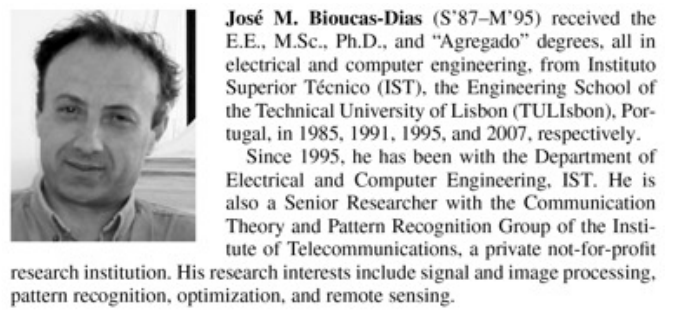}&
\includegraphics[width=\textwidth]{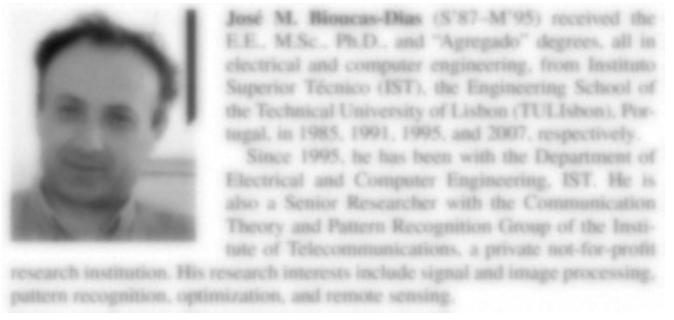}&
\includegraphics[width=\textwidth]{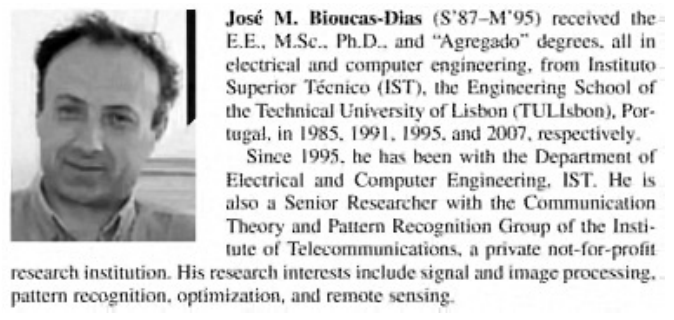}&
\includegraphics[width=\textwidth]{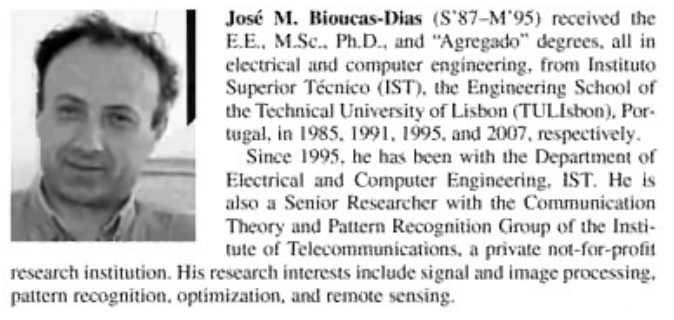}\\
(a)&(b)&(c)&(d)\\
\includegraphics[width=\textwidth]{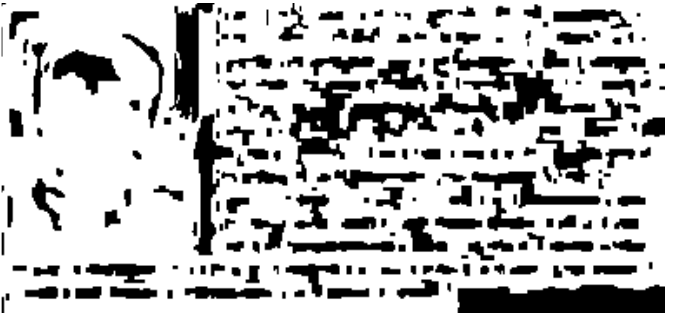}&
\includegraphics[width=\textwidth]{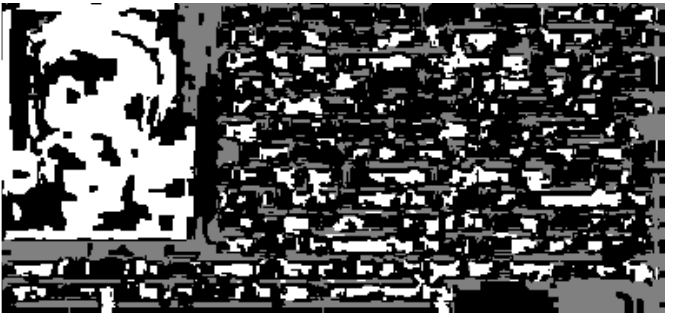}&
\includegraphics[width=\textwidth]{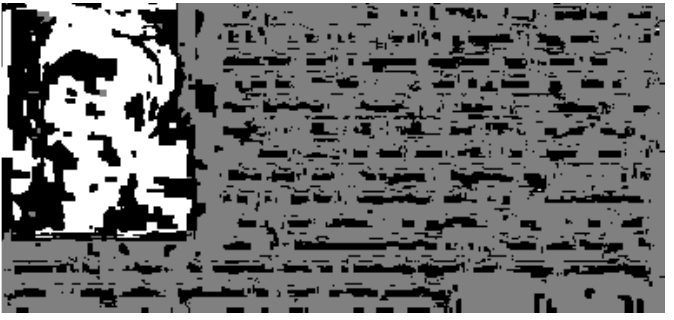}&
\includegraphics[width=\textwidth]{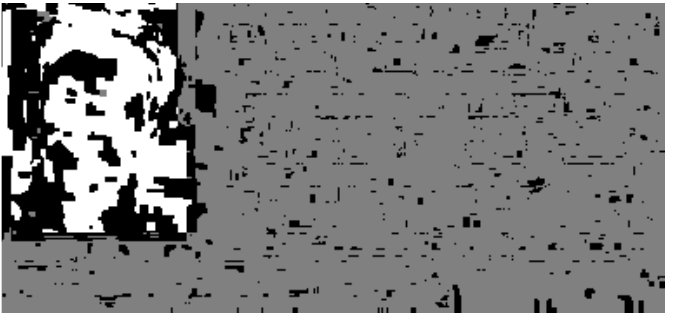}\\
(e)&(f)&(g)&(h)\\
\end{tabular}}
\caption{Deblurring: (a) input image Face + Text; (b) blurred image (Experiment 1); (c) deblurred with IDD-BM3D \cite{danielyan} (ISNR = 9.71dB); (d) deblurred with ADMM-GMM with $\alpha$-expansion (ISNR = 10.63dB); (e) $\alpha$-expansion (1st iteration); (f) $\alpha$-expansion (10th iteration); (g) $\alpha$-expansion (50th iteration); (h) $\alpha$-expansion (100th iteration). }
\label{fig:deb1}
\end{figure*}

Figure~\ref{fig:deb1} shows the patch labelling at different iterations. At first, as one would expect, the classes are not correctly identified; as the algorithm progresses, the labelling becomes more accurate. In this example we used six different classes but, for visualization purposes, we display only three: \textit{face} (white), \textit{text} (grey), and \textit{other} (black).


\section{Conclusions and Future work}
\label{sec:conclusions}

Recent work \cite{luo}, \cite{Teodoro2016} has shown that class-adapted image priors are able to outperform generic ones, when the input image in fact contains one or more of the considered classes. In this paper, we developed a method that automatically identifies which patches of the observed image should be reconstructed using class-adapted priors. This is an important feature for two main reasons: first, we often do not know whether the input image is from a particular class or not; second, we may have more than one class in a single image. 

Several aspects of the proposed approach still need improvement. First, although there is some very recent work regarding the convergence of ADMM with plug-and-play denoisers, there is a need to carefully analyse the convergence of ADMM with the GMM-based denoiser. Indeed, convergence of the algorithm is observed in practice, but theoretical support is still not available. Second, in the experiments reported in Section~\ref{sec:results}, parameter $\mu$ was hand-tuned; in future work, we will pursue more sophisticated techniques, such as \cite{AADMM}, to adjust this parameter in an automatic way.

Finally, in this paper we tested only a few very distinct image classes, but using more pre-computed models could eventually lead to better results. Some examples of possible classes are \textit{textures}, \textit{sky}, \textit{buildings}, and so on.

\end{document}